\documentclass{article}
\usepackage{spconf,amsmath,epsfig}
\usepackage{graphicx,lipsum}
\usepackage{tabularx}
\usepackage{multirow} 
\usepackage{anyfontsize}

\usepackage{booktabs}   
\usepackage{xcolor} 
\usepackage{multicol}
\usepackage{amsmath}
\usepackage{hyperref}
\hypersetup{
    colorlinks=false,
    }

\def\ourmethod{{Semantic Align Net (SAN)}}
\def\baseline{{Semantic Channel Net (SCN)}}
\def\ourmethodshort{{SAN}}
\def\baselineshort{{SCN}}

\title{A Semantic Segmentation-guided Approach for\\ Ground-to-Aerial Image Matching}
\name{Francesco Pro$^1$, Nikolaos Dionelis$^2$, Luca Maiano$^1$, Bertrand Le Saux$^2$, Irene Amerini$^1$} 
\address{$^1$Sapienza University of Rome, Italy
\\
$^2$European Space Agency (ESA), ESRIN, $\Phi$-lab, Italy}  
\begin{document}     
%
\maketitle          
\begin{abstract}  
Nowadays the accurate geo-localization of ground-view images has an important role across domains as diverse as journalism, forensics analysis, transports, and Earth Observation. This work addresses the problem of matching a query ground-view image with the corresponding satellite image without GPS data. This is done by comparing the features from a ground-view image and a satellite one, innovatively leveraging the corresponding latter's segmentation mask through a three-stream Siamese-like network. The proposed method, Semantic Align Net (SAN), focuses on limited Field-of-View (FoV) and ground panorama images (images with a FoV of 360°). The novelty lies in the fusion of satellite images in combination with their segmentation masks, aimed at ensuring that the model can extract useful features and focus on the significant parts of the images. This work shows how SAN through semantic analysis of images improves the performance on the unlabelled CVUSA dataset for all the tested FoVs.
\end{abstract}
\begin{keywords}   
Earth Observation data, Ground-to-aerial image matching, Semantic segmentation, Data fusion
\end{keywords}
\section{Introduction}
\label{sec:intro}
The accurate ground-to-aerial image matching task is a challenging problem to tackle, which raises high interest today. The correct geo-localization of ground view images in Computer Vision plays an important role across different domains, including journalism, forensics analysis, and Earth Observation. 
With the rapid increase of satellite images covering a huge part of the Earth’s surface,
the task of determining the geographical location represented in a given ground-view image, becomes a cross-view matching problem between the images taken by a camera placed at the ground level and the aerial images taken by a satellite or aerial vehicle, e.g. drone.

The different point of view between an image captured at ground level and that taken by a satellite poses an alignment problem between the two views. Added to this, the change of context and differences in terms of image quality between the two captured views add another level of complexity.
The ground images generally have a limited field of view, which is defined as the observable area captured by an optical device; the lower this value, the lower the quantity of information inside the image that is useful for comparison with an aerial one. 
A series of AI methods have been exploited in the last years to improve the image-matching problem. 
This work addresses the problem of matching a query ground-view image with the corresponding satellite image, proposing a method that focuses on both ground panoramas and limited \emph{Field-of-View} (FoV) image data.
\begin{figure}[tb]
\begin{minipage}[b]{1.0\linewidth} 
  \centering
  \centerline{\epsfig{figure=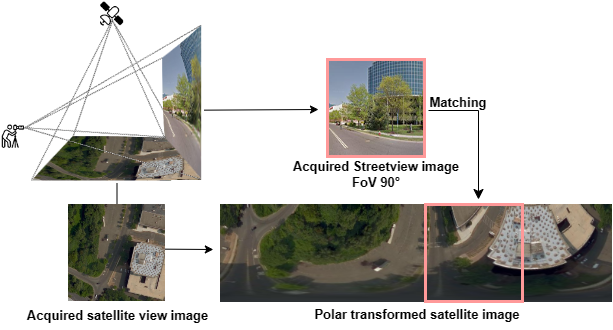,width=8.5cm}}
  \vspace{0cm}
\end{minipage}
\caption{Example of the ground-to-aerial matching problem. The \emph{query} ground-view image is matched to the polar transformed aerial image.}
\label{fig:res}
\end{figure}

The novelty of this work is in the introduction of satellite images in combination with their estimated segmentation masks, as additional information, to ensure that the model can focus on the significant parts of the images, such as roads, buildings, and vegetation that can be helpful for the data matching. 
In particular, our Semantic Align Net (SAN) model is based on a Siamese-like network \cite{shi2019optimal, shi2020i, Rodrigues_2022_WACV} with three CNN branches extracting features from the ground, the satellite, and the segmentation images. The last two mentioned extracted features are concatenated and, then, correlated with the ones extracted from ground images. 
The input data to this model are pre-processed, performing normalization, polar transformation, and semantic segmentation on labelled and unlabelled data.
The SAN model comprises a triplet loss that is used to train the networks using a similarity function, estimating the orientation angles between the images. 
Figure~\ref{fig:res} shows the points of view from which the two images are captured and the red box represents the identification of the query ground view image inside the satellite one. 
To overcome the scarcity of labelled aerial images and the difficulties of creating annotations for these images, we have implemented a new method for performing semantic segmentation from unlabelled data (NEOS) in parallel with this work~\cite{Dionelis2023}. This is a \textit{Transformer}-based model for semantic segmentation to segment satellite images of the non-annotated CVUSA dataset used for the geo-localization task. This allows us to evaluate our model on this unannotated dataset.
We show that the proposed ground-to-aerial matching model, SAN, improves the performance compared to the baseline models on a subset of the unlabelled CVUSA dataset on all the tested FoVs.

\section{Related Work}
\label{sec:related_works} 
The first attempts to this task came from ground-to-ground image-matching. Hays et al.~\cite{Hays:2008:im2gps} introduced a method aimed to geolocate through the extraction of any type of geographical information from ground-level images.
Instead, Chen et al.~\cite{5995610} focused on image relationships through 3D reconstruction and geometric constraints in urban and natural environments. Geometric constraints were also used by Baatz et al.~\cite{baatz} mainly for natural environments. Shan et al.~\cite{7035866} instead contributed to the task's direction, a pipeline with a viewpoint-dependent matching method to handle ground-to-aerial variant viewpoints, while other works~\cite{bonaventura, verde2020ground} model this task as a graph matching problem.

In recent years, a series of methods based on Artificial Intelligence have been introduced that improve performance in this task. In particular, we can group these techniques into the following three main groups:
\\
\textbf{Generative Adversarial Networks (GAN)}. This model and architecture is used to synthesize aerial images from ground-view image data, thus reducing the domain gap between the ground and aerial image domains. 
In the end, this information is combined or fused for the retrieval task~\cite{regmi2019bridging, deng2019using, toker2021coming}.   
\\
\textbf{Vision Transformers (ViT)}. More recent studies introduce ViT exploiting a self-attention mechanism working on temporal sequences of images, trying to model and capture temporal correlations that could help the image matching~\cite{9859992, zhang2022crossview}. 
\\
\textbf{Siamese-like networks}. This architecture's methods try to learn viewpoint-invariant discriminative characteristics from the images~\cite{shi2019optimal, shi2020i, Rodrigues_2022_WACV, 8578856}. This is done by passing ground and aerial-view images to CNNs that extract features from them. Our proposed model, SAN, falls into this last category.


%
%


\begin{figure*}[ht]   

\begin{minipage}[b]{1.0\linewidth}
  \centering
  \centerline{\epsfig{figure=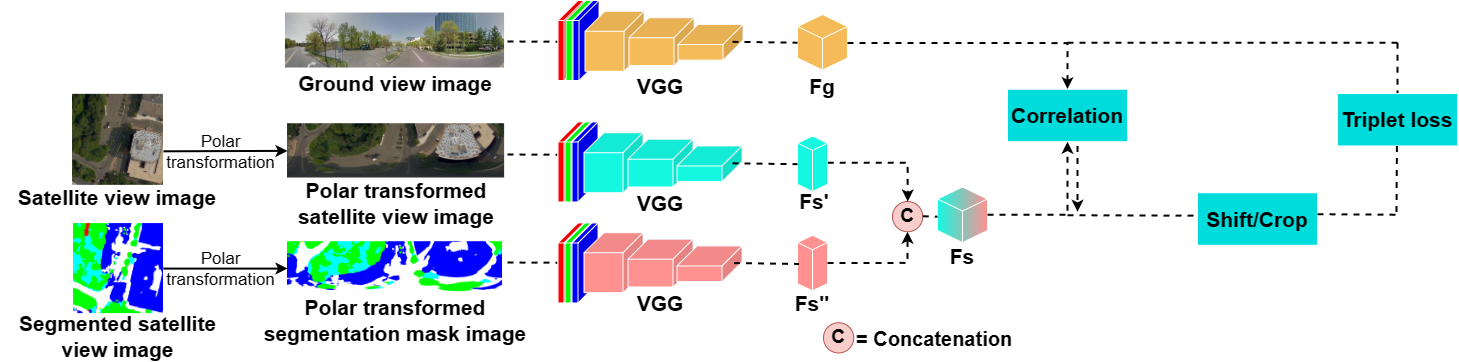,width=17cm,height=4.2cm}}
  \vspace{0cm} 
\end{minipage}
\caption{Our \emph{\ourmethodshort}\ network comprises three VGG16 branches extracting features from the (1) ground-view image, the (2) satellite-view image, and (3) its corresponding semantic segmentation mask. The features from the last two branches are correlated and compared with the ones from the ground image to estimate a similarity score.}  
\label{fig:model} 
\end{figure*}

\section{Proposed Method}
\label{sec:proposed} 

In this work, we propose a Siamese-like network, which we call \ourmethod, that combines features from aerial and ground-view images. Inspired by Shi et al.~\cite{shi2020i}, we apply the polar transformation to the satellite images to bring them into a ground domain format, reducing the gap between the aerial and ground domains. Then, the network is trained to estimate a similarity-matching score between them. Since ground-view images usually have different field-of-views, we compute this score at each possible orientation angle to augment the variability seen during training.

The main contribution of this work is the introduction of the satellite images in combination with their semantic segmentation masks, as additional information, hence making the model able to focus on the important parts of the images, such as on roads, buildings, and vegetation. 
In this way, the robustness to the image content variation is also increased.
In the next paragraphs, we will go into the details of the proposed model through the following steps: (1) Semantic Segmentation, (2) Image Processing, (3) Feature Extraction and Concatenation, and (4) Similarity Matching. Figure~\ref{fig:model} presents the proposed architecture.
\\
\textbf{Semantic segmentation.} In the Earth Observation (EO) field, it is very hard to find labelled aerial images, and it is also very difficult to annotate them \cite{castillonavarro2020semisupervised,chen_eo,7924536}. 
The CVUSA dataset used as benchmark for the ground-to-aerial image matching task contains non-segmented aerial images. 
Therefore, to extract the segmentations from aerial images, we rely on the Non-annotated Earth Observation Semantic Segmentation (NEOS)~\cite{Dionelis2023} method. This method is trained on labelled and unlabelled datasets, evaluating the unlabelled CVUSA dataset.
NEOS uses a Transformer-based model based on SegFormer~\cite{xie2021segformer}. It performs domain adaptation, minimizing the distribution differences of the different domains, so it can work on different datasets having distribution inconsistencies mainly due to acquisition scenes, environment conditions, sensors, and time.
\\
\textbf{Image pre-processing.} Ground-view and aerial images are captured from two completely different points of view, therefore introducing drastic changes of perspective between these two cross-view images. Therefore, in this step, we reduce the differences between the satellite and the ground/street images by cutting and resizing them to a specific shape before passing the images as inputs to the neural networks.
This domain gap is reduced by using the polar transformation, by applying the following two formulas:
\begin{equation}
x_i^s = \frac{D_s}{2} - \frac{D_s}{2}\ \frac{(H_v - x_i^t)}{H_v} \cos(\frac{2\pi}{W_v}y_i^t) \label{eq:special1}
\end{equation}
\begin{equation}
y_i^s = \frac{D_s}{2} + \frac{D_s}{2}\ \frac{(H_v - x_i^t)}{H_v} \sin(\frac{2\pi}{W_v}y_i^t) \label{eq:special2}
\end{equation}
where \(D_s \times D_s\) is the size of an aerial image, and \(H_v \times W_v\) denotes the target size of the polar transformation. 
Here, \((x_i^s,y_i^s)\) is a generic point on the original image, while \((x_i^t,y_i^t)\) is a generic point on the transformed image.
To ensure that the scale of the transformed images is consistent, the images in this work are first normalized by dividing by 255, and then centered with the dataset mean (\(\mu\)) and standardized with the dataset standard deviation (\(\sigma\)).
\\ 
\textbf{Feature extraction and concatenation.} The images are then passed to the networks. 
The proposed model, SAN, shown in Figure~\ref{fig:model}, is composed of three network branches: (1) one for the \emph{ground-view images}, (2) one for the \emph{satellite-view images}, and (3) one for the corresponding \emph{semantic segmentation mask images}.
Each branch uses a VGG16 model for extracting features and for generating the corresponding feature maps as output. 
Each of the networks in Figure~\ref{fig:model} is independent from the others, but they all have the same identical structure: (1) 13 Convolutional layers, (2) 3 Max Pooling layers, and (3) 3 Dropout layers.
The fully connected and classification layers are truncated from the architecture.

Depending on the chosen FoV, the ground network takes the input images and the output feature maps with variable widths as input. 
The satellite and segmentation mask network branches take both as input an image of shape (128x512x3) and output feature maps of shape (4x64x8). 
These two feature maps are concatenated on the third axis, obtaining a resulting shape of (4x64x16). 
The features are fused to weigh the important features extracted by the satellite images through the ones extracted by the semantic segmentation masks, thus providing a focus on the important elements inside the images.
\\
\textbf{Similarity matching.} In this step, the feature maps extracted from the images are correlated to obtain a similarity score between them and to estimate an orientation angle between polar transformed satellite images and ground-view images. 

In general, the relative orientation between the ground-view and the aerial images is not previously known, and it can differ among the images. Therefore, the polar transformation applied to the satellite images aligns them with the ground images up to an unknown azimuth angle.
Finding the correct alignment between the aerial and the ground images is crucial for accurate image matching, particularly for limited FoV ground-view images.
\begin{table*}[t]   
\large
\centering   
\resizebox{\textwidth}{!}{%
\begin{tabular}{|c|cccc|cccc|cccc|cccc|} 

\hline
                            Method & \multicolumn{4}{c|}{FoV 360° }                      & \multicolumn{4}{c|}{FoV 180°}                       & \multicolumn{4}{c|}{FoV 90°}                        & \multicolumn{4}{c|}{FoV 70°}                         \\ 
\cline{2-17}
                            & r@1 & r@5 & r@10 & r@1\% & r@1 & r@5 & r@10 & r@1\% & r@1 & r@5 & r@10 & r@1\% & r@1 & r@5 & r@10 & r@1\%  \\ 
\hline
\ourmethodshort\ (\emph{our})    & \textbf{77,07\%}      & \textbf{92,14\%}      & \textbf{95,62\%}       & \textbf{97,97\%}        & \textbf{48,49\%}      & \textbf{75,53\%}      & \textbf{84,06\%}       & \textbf{91,24\%}        & \textbf{6,23\%}       & \textbf{16,43\%}      & \textbf{24,20\%}       & \textbf{37,07\%}        & \textbf{3,02\%}       & \textbf{9,75\%}       & \textbf{15,44\%}       & \textbf{25,82\%}         \\
\baselineshort\ (\emph{our}) & 54,27\%      & 75.67\%      & 82,98\%       & 89,94\%        & 22,21\%            & 45,51\%            & 57,25\%             & 69,84\%              & 1,76\%            & 6,41\%            & 10,02\%             & 16,93\%              & 1,22\%            & 4,11\%            & 6,77\%             & 13,23\%               \\
Shi et al.~\cite{shi2020i}        & 72,87\%      & 89,62\%      & 92,78\%       & 96,52\%        & 45,28\%      & 72,87\%      & 81,76\%       & 89,57\%        & 6,14\%       & 15,80\%      & 22,93\%       & 35,17\%        & 2,21\%       & 7,31\%       & 12,60\%       & 20,50\%         \\
\hline
\end{tabular}%
}
\caption{Comparison between \ourmethodshort\ (proposed) and two baseline methods.\label{table:baselines}}
\end{table*}

\begin{table*}[t]
\large
\centering
\resizebox{\textwidth}{!}{%
\begin{tabular}{|c|l|llll|llll|llll|llll|} 
\hline
\multicolumn{1}{|c}{Trained Fov}                               &                     & \multicolumn{4}{c|}{Tested FoV 360° }                                                                                                             & \multicolumn{4}{c|}{Tested FoV 180°}                                                                                                              & \multicolumn{4}{c|}{Tested FoV 90°}                                                                                                               & \multicolumn{4}{c|}{Tested FoV 70°}                                                                                                                \\ 
\cline{3-18}
\multicolumn{1}{|l}{}                               &                     & \multicolumn{1}{c}{r@1}     & \multicolumn{1}{c}{r@5}     & \multicolumn{1}{c}{r@10}    & \multicolumn{1}{c|}{r@1\%}   & \multicolumn{1}{c}{r@1}     & \multicolumn{1}{c}{r@5}     & \multicolumn{1}{c}{r@10}    & \multicolumn{1}{c|}{r@1\%}   & \multicolumn{1}{c}{r@1}     & \multicolumn{1}{c}{r@5}     & \multicolumn{1}{c}{r@10}    & \multicolumn{1}{c|}{r@1\%}   & \multicolumn{1}{c}{r@1}     & \multicolumn{1}{c}{r@5}     & \multicolumn{1}{c}{r@10}    & \multicolumn{1}{c|}{r@1\%}    \\ 
\hline
\multirow{2}{*}{360°}        & \ourmethodshort\ (\emph{our})   & \multicolumn{1}{c}{\textbf{77,07\%}} & \multicolumn{1}{c}{\textbf{92,14\%}} & \multicolumn{1}{c}{\textbf{95,62\%}} & \multicolumn{1}{c|}{\textbf{97,97\%}} & \multicolumn{1}{c}{\textbf{47,63\%}} & \multicolumn{1}{c}{\textbf{75,30\%}} & \multicolumn{1}{c}{\textbf{83,43\%}} & \multicolumn{1}{c|}{\textbf{90,88\%}} & \multicolumn{1}{c}{\textbf{18,65\%}} & \multicolumn{1}{c}{\textbf{38,92\%}} & \multicolumn{1}{c}{\textbf{48,26\%}} & \multicolumn{1}{c|}{\textbf{60,00\%}} & \multicolumn{1}{c}{\textbf{12,28\%}} & \multicolumn{1}{c}{\textbf{28,04\%}} & \multicolumn{1}{c}{\textbf{36,84\%}} & \multicolumn{1}{c|}{\textbf{47,40\%}}  \\
                                                    & Shi et al.~\cite{shi2020i} & 72,87\%                              & 89,62\%                              & 92,78\%                              & 96,52\%                               & 44,47\%                              & 70,74\%                              & 79,01\%                              & 87,63\%                               & 18,10\%                              & 36,61\%                              & 45.60\%                              & 56,48\%                               & 11,87\%                              & 25,06\%                              & 33,00\%                              & 43,61\%                                \\ 
\hline
\multirow{2}{*}{180°} & \ourmethodshort\ (\emph{our})   & \multicolumn{1}{c}{\textbf{67,67\%}} & \multicolumn{1}{c}{\textbf{86,95\%}} & \multicolumn{1}{c}{\textbf{92,46\%}} & \multicolumn{1}{c|}{\textbf{96,12\%}} & \multicolumn{1}{c}{\textbf{48,49\%}} & \multicolumn{1}{c}{\textbf{75,53\%}} & \multicolumn{1}{c}{\textbf{84,06\%}} & \multicolumn{1}{c|}{\textbf{91,24\%}} & \multicolumn{1}{c}{\textbf{21,08\%}} & \multicolumn{1}{c}{\textbf{43,21\%}} & \multicolumn{1}{c}{\textbf{53,50\%}} & \multicolumn{1}{c|}{\textbf{65,91\%}} & \multicolumn{1}{c}{\textbf{14,67\%}} & \multicolumn{1}{c}{\textbf{30,84\%}} & \multicolumn{1}{c}{\textbf{39,82\%}} & \multicolumn{1}{c|}{\textbf{52,55\%}}  \\
                                                    & Shi et al.~\cite{shi2020i} & 66,64\%                              & 85,06\%                              & 91,11\%                              & 95,53\%                               & 45,28\%                              & 72,87\%                              & 81,76\%                              & 89,57\%                               & 20,77\%                              & 41,53\%                              & 52,42\%                              & 64,11\%                               & 13,23\%                              & 30,20\%                              & 39,77\%                              & 52,55\%                                \\ 
\hline
\multirow{2}{*}{90°}  & \ourmethodshort\ (\emph{our})   & \multicolumn{1}{c}{23,07\%}          & \multicolumn{1}{c}{\textbf{45,96\%}} & \multicolumn{1}{c}{\textbf{56,75\%}} & \multicolumn{1}{c|}{\textbf{70,16\%}} & \multicolumn{1}{c}{\textbf{13,72\%}} & \multicolumn{1}{c}{\textbf{30,97\%}} & \multicolumn{1}{c}{\textbf{42,80\%}} & \multicolumn{1}{c|}{\textbf{56,88\%}} & \multicolumn{1}{c}{\textbf{6,23\%}}  & \multicolumn{1}{c}{\textbf{16,43\%}} & \multicolumn{1}{c}{\textbf{24,20\%}} & \multicolumn{1}{c|}{\textbf{37,07\%}} & \multicolumn{1}{c}{\textbf{3,79\%}}  & \multicolumn{1}{c}{\textbf{12,10\%}} & \multicolumn{1}{c}{\textbf{18,87\%}} & \multicolumn{1}{c|}{\textbf{29,75\%}}  \\
                                                    & Shi et al.~\cite{shi2020i} & \textbf{23,48\%}                     & 43,57\%                              & 53,41\%                              & 66,59\%                               & 12,42\%                              & 29,21\%                              & 39,19\%                              & 54,99\%                               & 6,14\%                               & 15,80\%                              & 22,93\%                              & 35,17\%                               & 3,61\%                               & 11,47\%                              & 18,06\%                              & 28,04\%                                \\ 
\hline
\multirow{2}{*}{70°}  & \ourmethodshort\ (\emph{our})   & \textbf{14,67\%}                     & \textbf{33,18\%}                     & \textbf{42,93\%}                     & \textbf{56,52\%}                      & \textbf{7,27\%}                      & \textbf{21,40\%}                     & \textbf{30,43\%}                     & \textbf{44,20\%}                      & \textbf{3,43\%}                      & \textbf{12,19\%}                     & \textbf{19,82\%}                     & \textbf{31,24\%}                      & \textbf{3,02\%}                      & \textbf{9,75\%}                      & \textbf{15,44\%}                     & \textbf{25,82\%}                       \\
                                                    & Shi et al.~\cite{shi2020i} & 11,24\%                              & 26,37\%                              & 35,85\%                              & 49,93\%                               & 6,46\%                               & 15,58\%                              & 23,34\%                              & 36,07\%                               & 2,84\%                               & 9,57\%                               & 14,09\%                              & 24,74\%                               & 2,21\%                               & 7,31\%                               & 12,60\%                              & 20,50\%                                \\
\hline
\end{tabular}%
}
\caption{Generalization tests comparing \ourmethodshort\ (proposed) with the method by \emph{Shi et al.}~\cite{shi2020i}. \label{table:fovs}}
\end{table*}
To align with the correct orientation angle, for the ground-view and the aerial images, a rotation is needed, when reducing the domain gap between the two views. 
Aerial images are brought into the ground domain. 
At this point to find the correct orientation angle, we need a shift on the horizontal axis, both for the aerial and the ground-view images. 
Using this concept, the correlation between ground and satellite images can be computed along the azimuth angle by fixing the feature map \(F_s\) at the bottom and moving the feature map \(F_g\) on top of it as a sliding window, executing an inner product on each possible orientation angle. 
In this way, we obtain a similarity matching score for each orientation angle between the images.
For this purpose, we use the correlation operation between the two feature maps: $F_s \star F_g$. 
Let \(F_s\) \(\in R^{H\cdot  W_s\cdot C}\) and \(F_g\) \(\in R^{H \cdot W_v \cdot C}\) denote the aerial and ground features respectively, where H and C indicate the height and the channel
number of the features.
Here, \(W_s\) and \(W_v\) represent the width of
the aerial and ground features respectively. After the polar transformation, we have \(W_s\ \ge W_v\) for all images.
The correlation between \(F_s\) and \(F_g\) is expressed as: \
\begin{multline}  
  [F_s \star F_g](i) = \sum_{c=1}^{C}\sum_{h=1}^{H}\sum_{w=1}^{W_v} F_s(h, (i + w)\%W_s, c) \\ \cdot F_g(h, w, c) \label{eq:correlation}
\end{multline}where \(F(h, w, c)\) is the feature response at index \((h, w, c)\)\unboldmath, and \(\%\) denotes the modulo operation.
After this computation, the position of the maximum score corresponds to the estimated orientation between ground and aerial images. 
At the end of this operation, the two final feature maps are used to generate a matrix representing the similarity distance between each compared image. 
This operation is applied to both matching image pairs and non-matching images.
In this way, for matching image pairs, the network is forced to learn similar latent features, while for non-matching images, the similarity score is minimized once the orientation is computed so that the network can learn to discriminative features.

\section{Results}   
\label{sec:majhead}  
\textbf{Implementation details.} The three VGG16 networks are structured so that the first ten convolutional layers are pre-trained on the ImageNet dataset. The following three layers, instead, have been randomly initialized. The first seven layers are frozen, while the others are trainable.
The networks have been trained for 30 epochs using the Adam optimizer, with a starting learning rate of value \(10^{-5}\). Although other networks may fit our case, we will explore this study in the future.

Every experiment reported in this work is evaluated using the \emph{top-K recall} evaluation metric (in Table~\ref{table:baselines} and Table~\ref{table:fovs} is indicated with the notation r@1, r@5, r@10, and r@1\%), which measures how often a ground-view image is correctly matched with the corresponding aerial image, in terms of distance, among the first $K$ predicted output results. We also release our code for reproducibility\footnote{\url{https://pro1944191.github.io/SemanticAlignNet/}}. 
\\
\textbf{Datasets.} We train the semantic segmentation model~\cite{Dionelis2023} on the: (a) \emph{Potsdam}~\cite{potsdam}, and (b) \emph{Vaihingen}~\cite{vaihingen} datasets.
We also train and evaluate both our proposed ground-to-aerial image matching SAN method and~\cite{Dionelis2023} on the: (c) \emph{CVUSA}~\cite{workman2015widearea} dataset. 
The proposed method uses a filtered version of it~\cite{zhai2016predicting} 
The used ground view images have a FoV value of 360°, so as to work with different values. 
They are randomly cropped to have the desired width. 
For all the experiments, we train our model, SAN, and the baselines on $6647$ triplets of images and test on $2215$ images from~\cite{zhai2016predicting}. 
Here, a triplet comprises a ground-view image, a satellite-view image, and the corresponding segmentation mask, where all three represent the same location. 
%
\\
\textbf{Baselines.} We compare the performance of our model against two baselines. (1) The first is the state-of-the-art method introduced by Shi et al.~\cite{shi2020i}. (2) The second, called \baseline, is a two-branch network that we created as a baseline model to compare it with our SAN model. The first branch is for the ground-view images and the other one is obtained by concatenating the aerial images with their segmentation maks. Differently from our proposed \ourmethod~model, the concatenation is done at the image level and not at the features level.
\\
\textbf{Experiments.} For a reliable comparison, in this work, we use the same FoV values used by Shi et al.~\cite{shi2020i}: 360°, 180°, 90°, 70°. A different model is trained for each FoV value mentioned above. Based on this, we present two different types of experiments. In the first experiment, which is reported in Table~\ref{table:baselines}, we validate our model by comparing its performance with the two baseline methods introduced before. All three methods are trained and tested with the same CVUSA subset, using the same images split between the training set and the test dataset. 
According to the results, the proposed method improves the performance for all the FoV values. From Table~\ref{table:baselines}, we can observe a decrease in performance with lower FoV values, and this is mainly because of the fact that the lower the FoV value, the less information in the image.

Our second experiment, reported in Table~\ref{table:fovs}, is a generalization test in which a model trained on a specific FoV value is tested on other FoV values, i.e. on FoV values on which it was not trained. 
This experiment allows us to see how a model behaves with images with a different FoV from the ones it was trained on. Our model outperforms the one by Shi et al.~\cite{shi2020i} in all the experiments, thus showing that semantic segmentation introduces useful information to the model, which makes it more robust in all the FoV values.

\section{Conclusion}  
\label{sec:conclusion}   
This work examines geolocation without using GPS data, and to this end, we model the problem as a cross-view ground-to-aerial image matching task. 
Our main idea is to integrate the aerial view with its semantic segmentation mask in a three branches Siamese-like network. 
Our experiments show that this additional stream allows the network to focus on the correct parts of the aerial image and obtain more accurate image matching scores. 
Our proposed model, SAN, outperforms the other baseline models in all the experiments in this work.
Our future work will extend our experiments on the entire CVUSA dataset and will introduce the semantic segmentation masks for the ground-view image data.



\vfill
\pagebreak



\bibliographystyle{IEEEbib}
\bibliography{IGARSS/refs}

\end{document}